\documentclass[lettersize,journal]{IEEEtran}
\usepackage{amsmath,amsfonts}
\usepackage{algorithmic}
\usepackage{algorithm}
\usepackage{array}
\usepackage[caption=false,font=normalsize,labelfont=sf,textfont=sf]{subfig}
\usepackage{textcomp}
\usepackage{stfloats}
\usepackage{url}
\usepackage{verbatim}
\usepackage{graphicx}
\usepackage{cite}
\usepackage{booktabs}
\usepackage{makecell}

\begin{document}

\title{Genetic Programming with Model Driven Dimension Repair \\ for Learning Interpretable Appointment Scheduling Rules}

\author{Huan Zhang, Yang Wang, Ya-Hui Jia, \textit{Member, IEEE}, Yi Mei, \textit{Senior Member, IEEE},

\thanks{Huan Zhang and Yang Wang are with the School of Management, Northwestern Polytechnical University, Xi’an 710072, China. (email: zhhuan@mail.nwpu.edu.cn, yangw@nwpu.edu.cn).

Ya-Hui Jia is with School of Future Technology, South China University of Technology, Guangzhou 510641, China. (email: jia.yahui@foxmail.com).

Yi Mei is with School of Engineering and Computer Science, Victoria University of Wellington, Wellington 6140, New Zealand. (email: yi.mei@vuw.ac.nz).

This work is supported in part by the National Natural Science Foundation of China under Grant No. 72371200 and 71971172.}}

\markboth{IEEE Transaction of Evolutionary Computation,~Vol.~xx, No.~x,~xx~xxxx}
{Zhang \MakeLowercase{\textit{et al.}}: A New Dimensionally Aware Genetic Programming for Appointment Scheduling}


\maketitle

\begin{abstract}
Appointment scheduling is a great challenge in healthcare operations management. Appointment rules (AR) provide medical practitioners with a simple yet effective tool to determine patient appointment times. 
Genetic programming (GP) can be used to evolve ARs. However, directly applying GP to design ARs may lead to rules that are difficult for end-users to interpret and trust. A key reason is that GP is unaware of the dimensional consistency, which ensures that the evolved rules align with users' domain knowledge and intuitive understanding.
In this paper, we develop a new dimensionally aware GP algorithm with dimension repair to evolve ARs with dimensional consistency and high performance. A key innovation of our method is the dimension repair procedure, which optimizes the dimensional consistency of an expression tree while minimizing structural changes and ensuring that its output dimension meets the problem's requirements. We formulate the task as a mixed-integer linear programming model that can be efficiently solved using common mathematical programming methods. With the support of the dimension repair procedure, our method can explore a wider range of AR structures by temporarily breaking the dimensional consistency of individuals, and then restoring it without altering their overall structure, thereby identifying individuals with greater potential advantages. 
We evaluated the proposed method in a comprehensive set of simulated clinics. The experimental results demonstrate that our approach managed to evolve high-quality ARs that significantly outperform not only the manually designed ARs but also existing state-of-the-art dimensionally aware GP methods in terms of both objective values and dimensional consistency. In addition, we analyzed the semantics of the evolved ARs, providing insight into the design of more effective and interpretable ARs.
\end{abstract}

\begin{IEEEkeywords}
Appointment scheduling, dimensionally aware genetic programming, mixed-integer linear programming
\end{IEEEkeywords}

\section{Introduction}
\label{Introduction}

\IEEEPARstart{A}{ppointment} scheduling plays a crucial role in healthcare systems, impacting clinical, operational, and financial performance. A well-designed appointment system can improve the efficiency of medical providers and patient satisfaction by smoothing demand and mitigating uncertainty in patient arrivals \cite{zacharias2024dynamic}. 
Patients expect fast and timely service and find long wait times increasingly difficult to tolerate. Medical providers face pressure to efficiently utilize resources, including doctors' availability and expensive diagnostic machines. 
The overall goal of appointment scheduling is to achieve a balance between minimizing patients' waiting time and clinics' operational costs. Designing an effective appointment system has received considerable attention from academia \cite{cayirli2003outpatient, mondschein2003appointment, gupta2008review}.

During the past decades, the literature on appointment scheduling has focused primarily on developing the best appointment rules (AR) \cite{bailey1952study, cayirli2012universal, creemers2021evaluation}. AR offers medical providers a simple and practical decision-making tool to assign time slots to patient appointments. AR describes appointment intervals and the capacity of time slots within an appointment session using mathematical expressions \cite{cayirli2003outpatient}. 
In the early days, ARs were designed based on intuition, either assigning all patients to arrive at the start of an appointment session or distributing each patient a unique appointment time, evenly spaced throughout an appointment session \cite{soriano1966comparison}. 
Subsequent studies introduced more complex ARs, which were manually designed to suit specific clinic scenarios. It typically involved enumerating parameter values and evaluating their performance through extensive simulations, which is a time-consuming, trial-and-error process. To overcome these limitations, we adopt genetic programming to automate the design of ARs \cite{burke2009}. 

Genetic programming (GP) has shown great potential in symbolic regression \cite{vladislavleva2008order}, job-shop scheduling \cite{zhang2024}, vehicle routing \cite{wang2021genetic}, bin packing \cite{burke2006evolving} and other problems.
First, GP offers flexible representations based on symbolic expression trees, which can capture complex relationships between variables \cite{koza1992genetic}. Second, GP's powerful search mechanisms allow it to explore a wide range of program structures and find near-optimal solutions. Furthermore, the programs obtained by GP are partially interpretable, facilitating direct analysis of their semantics \cite{nguyen2017genetic}. 
Although the GP framework is highly generalizable, its components, such as the terminal set, function set, and fitness evaluation, must be tailored to each problem. To our knowledge, no studies have applied GP to the automatic design of ARs.

Unfortunately, directly applying standard GP does not always yield satisfactory results, as the generated rules are often semantically invalid.
This limitation arises because some terminals/features extracted from the problem correspond to physical measurements with specific units (e.g., mass, time and length), and the outputs of an AR must be appointment time. During the evolution process, GP randomly combines terminals and functions, which may result in an AR that performs well but lacks semantical interpretability and fail to produce valid appointment times. 
In contrast, although the manually designed ARs are less effective in terms of objective values, they ensure dimensional consistency and offer greater interpretability, which is crucial for real-world applications. 
Dimensionally aware genetic programming (DAGP) has been proposed to evolve rules with dimensional consistency. DAGP has been successfully applied to a variety of problems, such as machine discovery \cite{keijzer1999dimensionally}, empirical induction \cite{babovic2000genetic}, nonparametric modeling \cite{ratle2001grammar}, and automated innovation \cite{bandaru2013dimensionally}.
DAGP associates terminals with their measurement units and extends functions to operate in a dimension-aware manner, handling both mathematical and dimensional computations \cite{keijzer1999dimensionally}. This violates the closure property originally proposed by Koza, which requires all functions to be closed under any combination of functions and terminals \cite{koza1992genetic}. Once units are introduced, certain operations, such as adding a length to a time, become invalid or penalized. As a result, DAGP must be equipped with the ability to manage such invalid node combinations within the tree structure \cite{keijzer1999dimensionally}.

Existing DAGP methods can be categorized into two types: weakly typed DAGP \cite{keijzer1999dimensionally, keijzer2000genetic, mei2017constrained} and strongly typed DAGP \cite{ratle2001grammar, hunt2016evolving, da2021using}. 
Weakly typed DAGP allows dimensionally inconsistent combinations, but it manages the overall dimensional inconsistency of each individual and guides the population toward more dimensionally consistent models. Strongly typed DAGP defines the search space of GP using grammars and node types, restricting dimensionally inconsistent combinations during the initialization and breeding stages. 
Weakly typed DAGP can usually achieve better performance than strongly typed DAGP, but it cannot guarantee that its output always satisfies dimensional consistency. Although strongly typed DAGP can guarantee the dimensional consistency, the restrictions on node combinations make a large part of the search space infeasible, increasing the likelihood of the search getting trapped in poor local optima \cite{mei2017constrained}. 
Overall, current DAGP methods face a dilemma where dimensional consistency and performance cannot be achieved together. 

Motivated by the above considerations, we propose a novel DAGP algorithm to learn ARs with both high performance and dimensional consistency. The core idea is to allow genetic operators to freely combine nodes, thereby exploring a wider search space of AR. Dimensional consistency is then restored by selectively replacing a small number of terminals and functions, preserving the remaining structure of each individual's expression tree. 
Specifically, we develop a dimension repair procedure that optimizes the dimensional consistency in expression trees while minimizing structural changes, preserving their potentially effective patterns. 
This procedure ensures that the output measurement units of expression trees meet problem requirements and selects the most appropriate terminal replacements. 
We also incorporate an archive strategy that records elite individuals found during the evolutionary process and their corresponding terminal importance, to guide the dimensional repair procedure in selecting terminals for replacement. Furthermore, the archive also helps reduce the tree size
by merging itself with the current population to create a mating pool before breeding. The main contributions of this paper are summarized as follows:

\begin{enumerate}
\item {We propose a novel DAGP algorithm called GP with dimension repair (GPDR) to develop ARs with realistic semantic meaning.
We design a terminal set and an individual representation tailored to the problem. To the best of our knowledge, we are the first to introduce GP to solve the appointment scheduling problem.}

\item {We develop a dimension repair procedure to restore the dimensional consistency of an expression tree while minimizing its structural changes. We formulate the task of repairing the dimensional consistency of an expression tree as an optimization problem, and model it as a mixed integer linear programming (MILP) model that can be solved using common mathematical programming methods. This procedure enables GPDR to combine the exploratory power of weakly typed DAGP with the ability of strongly typed DAGP to maintain dimensional consistency among individuals.}

\item {We evaluate GPDR against manually designed AR and state-of-the-art DAGP methods across a wide range of simulated clinics. The results demonstrate that GPDR consistently produces high-performing and dimensionally consistent AR, outperforming other DAGP methods. Meanwhile, a semantic analysis of the evolved ARs is made to interpret their behavior.}
\end{enumerate}

The remainder of this paper is structured as follows. Section \ref{Background} provides an overview of the background on problem description, AR, and DAGP. Section \ref{Proposed Approach} details the GPDR algorithm and the proposed dimension repair procedure. Experimental studies and additional analysis are presented in Sections \ref{Experimental Studies} and \ref{Further Analysis}, respectively. Finally, Section \ref{Conclusion} concludes the paper.

\section{Background}
\label{Background}

\subsection{Problem Description}
\label{Problem description}

Appointment scheduling problem can be formally described as follows. Consider a single-server system, where $P$ patients need to be scheduled during an appointment session of duration $L$. Indices of patients are represented by $i=0,\dots,P-1$, sorted by their appointment times. Let $A_i$ denote the appointment time of patient $i$. 
The service times for these patients are modeled as independent and identically distributed random variables, following a log-normal distribution with a mean $M$, a standard deviation $V$, and a coefficient of variation $CV$. This assumption is based on empirical data in the literature \cite{cayirli2012universal}. To ensure that the capacity of the clinic matches the number of scheduled patients, the relationship $M \times P = L$ always holds. Appointment scheduling typically considers two environmental factors: no-shows and walk-ins. $PN$ is the probability that a scheduled patient does not show up, while $PW$ is the probability that a patient shows up without an appointment. The decision maker must establish an appointment timetable to efficiently schedule appointments of patients, taking into account these uncertainties. 

Performance measures for an appointment system include: 
$WAIT$ is patients' average waiting time, defined as the difference between the time their consultation actually begins and their appointment time for scheduled patients or arrival time for walk-in patients.
$IDLE$ is doctor's idle time per patient, defined as the period that the doctor remains unoccupied during the appointment session.
$OVER$ is doctor's overtime per patient, defined as the extra time required to serve all patients after the appointment session. 
The objective of appointment scheduling problem is to minimize the expected total cost of the system per patient, denoted as $TC$, which consists of two components: patient-related cost represented by $WAIT$ and doctor-related cost represented by a combination of $IDLE$ and $OVER$ with a fixed ratio of 1.5, as suggested by \cite{cayirli2012universal}. The relative importance of them can be adjusted based on the cost ratio between patient-related and doctor-related costs, referred to $CR$. Thus the total cost per patient $TC$ is expressed as:

\begin{equation}
\label{objective function}
TC= WAIT + CR \times (10 \times IDLE + 15 \times OVER)
\end{equation}

\subsection{Appointment Scheduling Rules}
\label{Appointment Scheduling Rules}

The literature on appointment system dsesign aims to find the best AR, which is a template to determine patient appointment times based on a specific combination of block sizes and appointment intervals \cite{cayirli2012universal}. Block size is the number of patients scheduled to the same time slot, who can be called individually or in groups. Appointment interval is the time between two consecutive appointments, which can be fixed or variable. One of the simplest AR is known as Individual Block/Fixed Interval, \textbf{IBFI} for short, which assigns each patient a unique appointment time, evenly spaced throughout an appointment session \cite{cayirli2003outpatient}.

The earliest study on AR dates back to Bailey \cite{bailey1952study}, who proposed an AR called \textbf{2BEG}. This rule schedules two patients at the start of the session, with the remaining patients singly at fixed intervals.
Soriano \cite{soriano1966comparison} studied the two-at-a-time appointment system and reported that a Multiple Block/Fixed Interval rule, referred to as \textbf{MBFI}. This rule assigns two or more patients to each appointment block and performs better in some of the environments studied.
Fries et al. studied appointment systems with variable block sizes and fixed intervals using a dynamic programming approach \cite{fries1981determination}.
Ho and Lau \cite{ho1992minimizing, ho1999evaluating} evaluated the performance of 50 ARs in a variety of environments and introduced an AR called \textbf{OFFSET} that allowed for variable intervals between two consecutive appointments.
Yang et al. \cite{yang1998new} proposed a universal AR for a single-server and multiple-customer service system. The new appointment rule was expressed as a mathematical function of environmental parameters, which determine its behavior.
Cayirli et al. \cite{cayirli2006designing} studied the interaction between AR and appointment sequencing rules and proposed an AR called \textbf{DOME}. This rule is designed based on the analytical studies, which identify a ``Dome" pattern in optimal solutions, where appointment intervals gradually increase toward the middle of the session and then decrease slightly towards the end. In subsequent research \cite{cayirli2008assessment}, they also studied the combination of AR and patient classification rules. They further proposed a universal AR in the presence of no-shows and walk-ins, as well as a universal AR considering patient heterogeneity \cite{cayirli2012universal, cayirli2014universal}.
The most recent research in this field was by Creemers et al. \cite{creemers2021evaluation}, who evaluated 314 ARs and found that a variant of 2BEG performed well across different environments. We refer to this as \textbf{RULE7}, as it was the seventh AR evaluated in this study.

Table \ref{Table_AR} shows the representative ARs, which have been highlighted in bold in the paragraphs above.
To calculate $A_i$, these rules typically use $M$ and $i$ to determine the appointment interval while introducing dynamic offsets based on $V$ and $i$ to simulate the variations of appointment intervals. Most importantly, the structures of manually designed AR always maintain dimensional consistency. Note that $A_i$, $M$, and $V$ are all physical measurements of time, while $i$ and other parameters are dimensionless. 

\begin{table}[!htp]\centering
\caption{Representative appointment rules}
\setlength{\tabcolsep}{5mm}{
\begin{tabular}{ll}

\toprule
AR & Expressions \\
\midrule
IBFI & $A_i = iM$ \\
\specialrule{0.1pt}{0pt}{0pt}
2BEG & \makecell[l]{ $A_i = 0, \text{ if } i \leq 1$  \\  
       $A_i = (i-1)M, \text{ if } i > 1$ } \\
\specialrule{0.1pt}{0pt}{0pt}
MBFI & \makecell[l]{ $A_i = iM, \text{ if } i = 0,2,4,\dots$ \\
       $A_i = (i-1)M, \text{ if } i = 1,3,5,\dots$ } \\
\specialrule{0.1pt}{0pt}{0pt}
OFFSET & \makecell[l]{ $A_i = iM + 0.15(i-k)V, \text{ if } i \leq k$ \\
         $A_i = iM + 0.3(i-k)V, \text{ if } i > k$ } \\
\specialrule{0.1pt}{0pt}{0pt}
DOME & \makecell[l]{ $A_i = iM + 0.15(i-k_1)V, \text{ if } i \leq k_1$ \\
       $A_i = iM + 0.3(i-k_1)V, \text{ if } k_1<i \leq k_2$ \\
       $A_i = iM - 0.05(i-k_2)V, \text{ if } i > k_2$ } \\
\specialrule{0.1pt}{0pt}{0pt}
RULE7 & \makecell[l]{ $A_i = 0, \text{ if } i \leq 1$ \\
        $A_i = (i-1)M + 0.3(i-1)V, \text{ if } i > 1$ } \\
\bottomrule

\end{tabular}}
\small{Note: \( k, k_1, k_2 \) are a parameter related to the number of patients \( N \).}
\label{Table_AR}
\end{table}

\subsection{Dimensionally Aware Genetic Programming}
\label{Dimensionally Aware Genetic Programming}

DAGP leverages both the numerical values of terminals and their corresponding measurement units. In DAGP, functions and terminals must be augmented to preserve the information of the measurement units they use, as shown below. 

\subsubsection{Definition of terminals}
\label{Definition of terminals}

Each variable and constant in the terminal set is associated with a dimension vector that represents the exponents of its physical measurement units. Traditionally, this vector is expressed as a vector $\mathbf{d} = [l, t, m]$, where $l$, $t$, and $m$ denote the exponents of the fundamental dimensions of length, time, and mass, respectively. For example, the dimension vector of acceleration (with its measurement unit of $m/s^2$) is $[1, -2, 0]$, while the dimension vector of a constant is $[0, 0, 0]$.


\subsubsection{Definition of functions}
\label{Definition of functions}

In addition to performing mathematical and logical operations, each function must also support dimensional operations. This includes clearly defining how measurement units are transformed and ensuring that all unit-related constraints are satisfied during the function application. 
Table \ref{Table_Functions} summarizes the effects of common functions used in GP on measurement units. The operations of addition, subtraction, max, and min are subject to the constraint that their operands and output must have the same dimension vectors. Multiplication and division combine their operands by adding and subtracting the dimension vectors, respectively. Square and square root multiply and divide the dimension vector by 2, respectively. Finally, the if-then-else operation places no constraint on the dimension of the first operand. However, the second and third operands, as well as the output, must share the same dimension vectors.

\begin{table}[!htp]\centering
\caption{Dimensional Operations of Functions}
\setlength{\tabcolsep}{3mm}{
\begin{tabular}{lccl}

\toprule
Function & Input Dimensions & Output Dimension\\ 
\midrule
Add  & $\mathbf{d_1}$, $\mathbf{d_2}$ & $\mathbf{d} = \mathbf{d_1} = \mathbf{d_2}$ \\
Sub  & $\mathbf{d_1}$, $\mathbf{d_2}$ & $\mathbf{d} = \mathbf{d_1} = \mathbf{d_2}$ \\
Max  & $\mathbf{d_1}$, $\mathbf{d_2}$ & $\mathbf{d} = \mathbf{d_1} = \mathbf{d_2}$ \\
Min  & $\mathbf{d_1}$, $\mathbf{d_2}$ & $\mathbf{d} = \mathbf{d_1} = \mathbf{d_2}$ \\
Mul  & $\mathbf{d_1}$, $\mathbf{d_2}$ & $\mathbf{d} = \mathbf{d_1} + \mathbf{d_2}$ \\
Div  & $\mathbf{d_1}$, $\mathbf{d_2}$ & $\mathbf{d} = \mathbf{d_1} - \mathbf{d_2}$ \\
Sq   & $\mathbf{d_1}$ & $\mathbf{d} = 2 \mathbf{d_1}$ \\
Sqrt & $\mathbf{d_1}$ & $\mathbf{d} = \mathbf{d_1} / 2$ \\
If-then-else & $\mathbf{d_1}$, $\mathbf{d_2}$, $\mathbf{d_3}$ & $\mathbf{d} = \mathbf{d_2} = \mathbf{d_3}$ \\
\bottomrule

\end{tabular}}
\label{Table_Functions}
\end{table}

These dimensionally augmented terminals and functions designed for DAGP violate the closure property. The closure property requires that each function in the function set must be well-defined for any combination of arguments from the function set and the terminal set \cite{koza1992genetic}. To address this challenge, a series of strategies have been proposed. 

One stream in the literature focuses on weakly typed DAGP.
Keijzer and Babovic \cite{keijzer1999dimensionally, babovic2000genetic} proposed a selection pressure-based DAGP that introduces an extra objective to measure the dimensional inconsistency of individuals.
They developed a culling function based on the total dimensional inconsistency of each individual. Crossover or mutation operators will generate multiple offspring, and only the one with the least dimensional inconsistency is added to the intermediate population. They also adopted a multi-objective optimization routine to find the pareto front of non-dominated individuals, balancing their performance and dimensional consistency \cite{keijzer2000genetic, babovic2001evolutionary}. 
Bandaru and Deb \cite{bandaru2013dimensionally} made the GP system dimensionally aware by introducing penalties for operations performed between incommensurable quantities.
Mei et al. \cite{mei2017constrained} developed a constrained DAGP that employs a constrained fitness function, defined by both the objective value and dimensional inconsistency. They designed a penalty coefficient adaptation mechanism to achieve a balance between performance and dimensional inconsistency during the GP search process. 
Li and Zhong \cite{li2020dimensionally} used multi-objective DAGP to study automatic crowd behavior modeling.
Weakly typed DAGP does not enforce strict dimensional consistency but adds selection pressure to promote overall dimensional consistency within the population. This means that its search space is very large and includes many irrelevant individuals \cite{ratle2001grammar}. It may require multiple runs to obtain a satisfying result.

Another stream in the literature studies strongly typed DAGP.
Ratle et al. \cite{ratle2000genetic, ratle2001grammar} present a grammar-guided GP that enforces dimensional constraints using a context-free grammar within the GP framework. They also propose an initialization procedure based on dynamic grammar pruning to generate individuals that respect the maximum tree depth. 
Silva \cite{da2021using} used grammar-guided GP to find dimensionally consistent dispatching rules for job-shop scheduling. 
In grammar-guided GP, each individual is represented as a derivation tree derived from a context-free grammar and needs to be translated into a conventional expression tree. 
The application of grammar-guided GP faces several limitations.
Firstly, it only considers integer exponents of measurement units restricted to the range $\{ -2, -1, 0, 1, 2 \}$. For a dimension vector with a norm of 3, this means the non-terminal symbols in the grammar must be defined over a domain of $5^3 = 125$ possible combinations. This results in a highly complex grammar, often necessitating the use of automated grammar generators for its construction \cite{ratle2000genetic}.
Secondly, such complex grammars often fall under the category of explosive grammars, where the probability of adding a non-terminal symbol during derivation is higher than that of adding a terminal symbol \cite{nicolau2018understanding}. As a result, the initialization process tends to generate very deep trees \cite{ratle2001grammar}.
Finally, to prevent uncontrolled tree bloat, it is necessary to manually adjust the grammar to achieve a balanced grammar. Even small changes in grammar design have a significant impact on search \cite{dick2022initialisation}. 
Hunt et al. \cite{hunt2016evolving} implemented dimensional constraints using strongly typed GP to address the dynamic job-shop scheduling problem. However, their approach only considers exponents of measurement units as -1, 0, and 1, which limits their function set from using functions such as square and square root.
In summary, strongly typed GP ensures dimensional consistency in its output but treats continuous dimensions as discrete values, thereby excluding many potentially valid rule structures. Moreover, restrictions on node combinations in strongly typed GP also affect its search capacity, as its genetic operators are likely to fail to create structurally valid expressions \cite{babovic2000genetic}. This results in the performance of strongly typed DAGP being inferior to that of weakly typed DAGP.

\section{Proposed Approach}
\label{Proposed Approach}

In this section, we first introduce our newly proposed GPDR algorithm and provide a detailed description of its individual components, including individual representation and evaluation, dimension repair procedure, archive update, and genetic operators.

\subsection{GPDR Framework}
\label{Overall Framework}

\begin{figure}\centering
\includegraphics[width=2.5in]{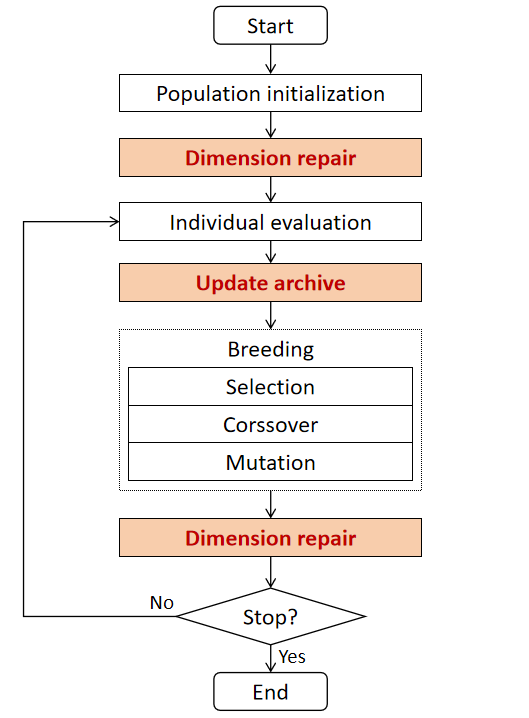}
\caption{The flowchart of the proposed GPDR algorithm.}
\label{Fig_Overall Framework}
\end{figure}

Figure \ref{Fig_Overall Framework} illustrates the overall framework of GPDR. It starts with initializing a population of individuals with the ramped half-and-half method \cite{koza1992genetic}. Each individual in the population represents an AR, defining how patient appointments should be arranged within a session. The dimension repair procedure is then applied to each individual to restore their dimensional consistency. 
In each generation, every individual is evaluated on a large number of simulation replications by calculating the average $TC$ of the appointment schedule it generates. 
GPDR maintains an archive throughout the search process, which stores non-dominated individuals identified so far based on fitness and program size. The archive is updated with any newly identified non-dominated individuals after each evaluation. 
The importance of each terminal is then calculated based on its frequency of occurrence in individuals within the archive. This information is used to guide node replacement in the subsequent dimension repair procedure.
In the breeding process, the archive is first merged with the current population, creating a mating pool. This combined pool guides the evolution of the population toward more compact and higher-quality solutions. Next, tournament selection is used to choose parents from the pool, followed by the application of crossover and mutation operators to generate offspring. After that, the dimension repair procedure is applied once again to compensate for the damage that genetic operators may cause to the dimensional consistency of individuals. Once the maximum number of generations is reached, GPDR terminates and returns the individual with the highest fitness from the archive. 

\subsection{Individual Representation and Evaluation}
\label{Individual Representation and Evaluation}

\begin{figure}\centering
\includegraphics[width=2.3in]{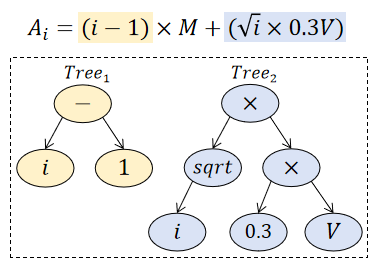}
\caption{An example of individual representation.}
\label{Fig_individual representationk}
\end{figure}

In this research, each individual is a tree-based expression representing an AR, dictating appointment times for each patient during an appointment session. Based on the observations summarized in Table \ref{Table_AR}, manually designed AR typically include a step size based on $M$, along with a correction term to adjust appointment intervals. To capture the structured patterns present in these rules, we propose a multi-tree individual representation:
\begin{equation}
A_i=Tree_1 \times M + Tree_2
\end{equation}
While physical systems may involve multiple dimensions, in our context of appointment scheduling, only time is relevant. Therefore, we simplify the dimension vectors by considering only the time component. Each variable or constant in the terminal set is thus assigned a scalar value $d$, representing its exponent with respect to the unit of time.
Since the dimension of $M$ is 1, $Tree_1$ must be an expression tree with a target dimension of 0, while $Tree_2$ is an expression tree with a target dimension of 1. The outputs of these two trees are then summed to obtain the appointment time, whose measurement unit has a dimension of 1.
Fig. \ref{Fig_individual representationk} illustrates an example of individual representation. Each individual consists of two expression trees, each tree representing a component of the mathematical expression. Moreover, by using a multi-tree representation, we reduce the size of each expression tree, which makes it less computationally expensive to restore dimensional consistency for each individual. To ensure the feasibility of the appointment schedule, we define the following rules for computing AR:
\begin{enumerate}
    \item If $A_i < 0$, set $A_i = 0$;
    \item If $A_i < A_{i-1}$, set $A_i = A_{i-1}$;
    \item If $A_i > L$, set $A_i = A_{i-1}$.
\end{enumerate}

The pseudocode of individual evaluation on a simulated clinic is outlined in Algorithm \ref{Individual Evaluation}. Each AR constructs an appointment schedules by assigning appointment time to each patient, starting from time zero. (line 3). The evaluation of individuals involves assessing the performance of these schedules through a set of simulation replications $R$.
In each simulation replication $r \in R$, scheduled patients are assumed to show up if a randomly generated number exceeds their no-show probability. Patients with appointments arrive punctually and are served on a First-In-First-Out basis, with only one patient being served at a time (lines 8-10). Walk-in patients are given the lowest priority but are guaranteed to be seen before more than three scheduled patients are served, following the policy in \cite{cayirli2006designing} (lines 11-13). After each replication, the total cost $TC_r$ is recorded (line 15). The fitness of an individual is then computed as the average total cost across all replications (line 17). A lower fitness indicates a more effective AR.
Further details regarding the simulation settings are provided in Section \ref{Simulation Model}. 
Preliminary experiments indicate that running 15,000 simulation replications per evaluation yields an average $TC$ within an error margin of 1\% at a confidence interval of 95\%, which is sufficient to reflect the true performance of an individual.

\begin{algorithm}
\caption{Individual Evaluation}
\label{Individual Evaluation}
\begin{algorithmic}[1]
    \STATE \textbf{Input}: Individual $Ind$
    \STATE \textbf{Output}: Fitness $f$
    \STATE Compute the appointment schedule for $Ind$
    \FOR{each $r \in R$} 
        \STATE // Do simulation
        \STATE Initialize the queue of appointment patients, excluding no-shows
        \WHILE{patients remain or clinic is not closed}
            \IF{doctor is idle}
                \STATE Serve next patient in the queue
            \ENDIF
            \IF{a walk-in patient arrives}
                \STATE Insert the walk-in patient at the 4\textsuperscript{th} position in the queue
            \ENDIF
        \ENDWHILE
        \STATE Compute the total cost $TC_r$ of current replication $r$ using Equation (\ref{objective function})
    \ENDFOR
    \STATE Compute fitness: $f = \sum_{r \in R} TC_r / |R|$
    \RETURN $f$
\end{algorithmic}
\end{algorithm}

\subsection{Dimension Repair Procedure}
\label{Dimension Repair Operator}

Dimension repair procedure is designed to optimize the dimensional consistency of an expression tree while minimizing its structural changes and ensuring that the output measurement units of expression trees meet problem requirements. The task of repairing the dimensional consistency of an expression tree can be described as an optimization problem and be formulated as an MILP model, as shown below.

\subsubsection{MILP Model}
\label{MILP Model}

Consider a symbolic expression tree consisting of a set of nodes denoted by $N$. Each node $n \in N$ belongs to one of two categories: terminal nodes represented by the set $N^t$, and function nodes represented by the set $N^f$. Together, these sets form the complete set of nodes: $N = N^t \cup N^f$. For each node $n \in N$, let the ordered array $\mathbf{c}_n$ represent its child nodes. If $n \in N^t$, then it has no child, and thus $||\mathbf{c}_n|| = 0$. If $n \in N^f$, then it may have one or more children, each of which is a terminal or a function. 

In the GPDR setup, each node $n \in N$ is associated with a dimension vector, denoted by $\mathbf{d}_n \in \mathbb{R}^D$. Each element of $\mathbf{d}_n$ is represented as $\mathbf{d}_{n,k}$ where $k$ indexes the dimension.
For terminal nodes, their dimension vectors are determined by the terminals themselves. For function nodes, their dimension vectors are determined by both the dimension vectors of their child nodes and their own dimension operations. Let $n*$ denote the root node of the tree and let $\mathbf{d}*$ represent the target dimension vector of the entire tree. 

To reduce the complexity of the model, we categorize the terminals and functions in DAGP into dimensionally equivalent classes.
Let $T$ be the set of terminal classes, where each $t \in T$ represents a class of terminals that share the same dimension vector. Let $\mathbf{v}_t \in \mathbb{R}^D$ denote the dimension vector of terminal class $t$. For example, the variables $M$ and $V$, both having the dimension vector $[0, 1, 0]$, belong to the same class of terminals.
Let $F$ be the set of function classes, where each $f \in F$ represents a class of functions that share the same dimensional constraints and operations. For example, functions such as addition, subtraction, max and min all have the same dimensional operation and thus belong to the same class of functions, as shown in Table \ref{Table_Functions}. Replacing terminals or functions with others from the same class does not affect the dimensional consistency of the tree. 

For the node set $N$, let $t_n$ denote the terminal class of each node $n \in N^t$, and let $f_n$ denote the function class of each node $n \in N^f$. The goal of the dimension repair procedure is to replace terminal and function nodes in the tree with nodes from other classes preserving the overall structure of the tree. To maintain this structure, functions can only be replaced by others that require the same number of child nodes. Let $F_f \subset F$ denote the set of function classes that can replace $f$. Table \ref{Table_Functions_types} summarizes all the function classes from the function set of GPDR and their compatible function classes. 

\begin{table}[!htp]\centering
\caption{Function classes based on dimensional operations}
\setlength{\tabcolsep}{5mm}{
\begin{tabular}{clll}

\toprule
$f$ & Functions &  $F_f$ \\ 
\midrule
1 & Add, sub, max, and min & 1, 2, 3 \\

2 & Mul & 1, 2, 3 \\

3 & Div & 1, 2, 3 \\

4 & Sq & 4, 5 \\

5 & Sqrt & 4, 5 \\

6 & If-else-then & 6 \\
\bottomrule

\end{tabular}}
\label{Table_Functions_types}
\end{table}

To restore the dimensional consistency of the tree, we introduce three binary variables to represent the modifications made to the nodes in the tree.

\begin{flalign}\label{constr2}
& x_n \text{ }\text{ } = 
\begin{cases}
  1, \text{if node } n \text{ requires adjustment;} \\ 
  0, \text{otherwise}\\
\end{cases} & 
\end{flalign}

\begin{flalign}
& y_{n, f} = 
\begin{cases}
  1, \text{if function node } n \text{ should be replaced} \\  \text{with a function in class } f;\\ 
  0, \text{otherwise}\\
\end{cases} & 
\end{flalign}

\begin{flalign}
& z_{n, t} = 
\begin{cases}
  1, \text{if terminal node } n \text{ should be replaced} \\  \text{with a terminal in class } t;\\ 
  0, \text{otherwise}\\
\end{cases} & 
\end{flalign}

These variables must satisfy the following constraints:

\begin{enumerate}
\item {Each function node only has one function class.}
\begin{equation}
\sum_{f\in F} y_{n,f} = 1, \quad \forall n \in N^f
\end{equation}

\item {Each function node can only be modified to a valid function class.}
\begin{equation}
y_{n,f} = 0,  \quad \forall n \in N^f , \forall f \in F / F_n
\end{equation}

\item {Each terminal node only has one terminal class.}
\begin{equation}
\sum_{t\in T} z_{n,t} = 1, \quad \forall n \in N^t
\end{equation}

\item {If a function node $n$ requires no adjustment ($x_n=0$), it cannot be modified.}
\begin{equation}
x_n + y_{n, f_n} \geq 1, \quad \forall n \in N^f
\end{equation}

\item {If a terminal node $n$ requires no adjustment ($x_n=0$), it cannot be modified.}
\begin{equation}
x_n + z_{n, t_n} \geq 1, \quad \forall n \in N^t
\end{equation}

\item {The dimension vector of a terminal node is determined by the terminal class after modification.}
\begin{equation}
\mathbf{d}_n = \sum_{t \in T} z_{n,t}\mathbf{v}_t, \quad \forall n \in N^t
\end{equation}

\item {The dimension vector of a function node is determined by its function class and child nodes after modification. If it should be replaced with function class 1, which is indicated by $y_{n,1} = 1$,
\begin{gather}
\mathbf{d}_{n,k} - \mathbf{d}_{n_1,k} \leq M(1 - y_{n,1}), \notag \\
\mathbf{d}_{n_1,k} - \mathbf{d}_{n,k} \leq M(1 - y_{n,1}), \notag \\
\forall n \in N^f, \forall k \in \{1, \dots, D\}, f_n \in F_1, \mathbf{c}_n = [n_1,n_2]
\end{gather}

\begin{gather}
\mathbf{d}_{n,k} - \mathbf{d}_{n_2,k} \leq M(1 - y_{n,1}), \notag \\
\mathbf{d}_{n_2,k} - \mathbf{d}_{n,k} \leq M(1 - y_{n,1}), \notag \\
\forall n \in N^f, \forall k \in \{1, \dots, D\}, f_n \in F_1, \mathbf{c}_n = [n_1,n_2]
\end{gather}

If it should be replaced with function class 2, which is indicated by $y_{n,2} = 1$,
\begin{gather}
\mathbf{d}_{n,k} - (\mathbf{d}_{n_1,k} + \mathbf{d}_{n_2,k}) \leq M(1 - y_{n,2}), \notag \\
(\mathbf{d}_{n_1,k} + \mathbf{d}_{n_2,k}) - \mathbf{d}_{n,k} \leq M(1 - y_{n,2}), \notag \\
\forall n \in N^f, \forall k \in \{1, \dots, D\}, f_n \in F_2, \mathbf{c}_n = [n_1,n_2]
\end{gather}

If it should be replaced with function class 3, which is indicated by $y_{n,3} = 1$,
\begin{gather}
\mathbf{d}_{n,k} - (\mathbf{d}_{n_1,k} - \mathbf{d}_{n_2,k}) \leq M(1 - y_{n,3}), \notag \\
(\mathbf{d}_{n_1,k} - \mathbf{d}_{n_2,k}) - \mathbf{d}_{n,k} \leq M(1 - y_{n,3}), \notag \\
\forall n \in N^f, \forall k \in \{1, \dots, D\}, f_n \in F_3, \mathbf{c}_n = [n_1,n_2]
\end{gather}

If it should be replaced with function class 4, which is indicated by $y_{n,4} = 1$,
\begin{gather}
\mathbf{d}_{n,k} - 2\mathbf{d}_{n_1,k} \leq M(1 - y_{n,4}), \notag \\
2\mathbf{d}_{n_1,k} - \mathbf{d}_{n,k} \leq M(1 - y_{n,4}), \notag \\
\forall n \in N^f, \forall k \in \{1, \dots, D\}, f_n \in F_4, \mathbf{c}_n = [n_1]
\end{gather}

If it should be replaced with function class 5, which is indicated by $y_{n,5} = 1$,
\begin{gather}
\mathbf{d}_{n,k} - 0.5\mathbf{d}_{n_1,k} \leq M(1 - y_{n,5}), \notag \\
0.5\mathbf{d}_{n_1,k} - \mathbf{d}_{n,k} \leq M(1 - y_{n,5}), \notag \\
\forall n \in N^f, \forall k \in \{1, \dots, D\}, f_n \in F_5, \mathbf{c}_n = [n_1]
\end{gather}

If it should be replaced with function class 6, which is indicated by $y_{n,6} = 1$,
\begin{gather}
\mathbf{d}_{n,k} - \mathbf{d}_{n_2,k} \leq M(1 - y_{n,6}), \notag \\
\mathbf{d}_{n_2,k} - \mathbf{d}_{n,k} \leq M(1 - y_{n,6}), \notag \\
\forall n \in N^f, \forall k \in \{1, \dots, D\}, f_n \in F_6, \mathbf{c}_n = [n_1,n_2,n_3]
\end{gather}

\begin{gather}
\mathbf{d}_{n,k} - \mathbf{d}_{n_3,k} \leq M(1 - y_{n,6}), \notag \\
\mathbf{d}_{n_3,k} - \mathbf{d}_{n,k} \leq M(1 - y_{n,6}), \notag \\
\forall n \in N^f, \forall k \in \{1, \dots, D\}, f_n \in F_6, \mathbf{c}_n = [n_1,n_2,n_3]
\end{gather}}

\item {The dimension vector of the root node must equal to the target output dimension vector of the tree.}
\begin{equation}\label{constr19}
\mathbf{d}_{n*} = \mathbf{d}*
\end{equation}

\end{enumerate}

Our goal is to minimize structural changes to the tree. For each node $n \in N$, we define a modification weight $w_n$ based on its depth. In GP, a node’s importance is closely correlated with its depth: deeper nodes are generally less important than those closer to the root \cite{luke2003modification}. Therefore, we assign weights as the inverse of the node's depth to reflect their importance. Specifically, the depth of the root node is defined as 1 to avoid division by zero. Based on the decision variables, constraints, and objective function described above, we formulate the dimension repair problem as the following MILP model:
\begin{flalign*}
\quad \min \quad & \sum_{n \in N} w_nx_n && \\
\quad \text{s.t.:} \quad & (2)\text{--}(19) &&
\end{flalign*}
Despite the problem is generally NP-hard, this model remains tractable in practice due to the limited size of evolved trees and the sparsity of feasible modifications.

\subsubsection{Node Replacement Strategy}
\label{Node Replacement Strategy}

By solving the MILP model, we have identified how to restore the dimensional consistency of the tree at minimal cost by replacing a few terminals and functions. However, selecting new nodes from terminal or function classes that are dimensionally equivalent remains a challenge.

For function replacement, if the new function belongs to classes 2 through 6, the replacement is straightforward, as each class has only one function. However, if the new function belongs to class 1, there are four possible functions in that class. The replacement rules are as follows: if the old function is from class 1, it remains unchanged; if the old function is from class 2, the new function will be an addition; if the old function is from class 3, the new function will be a subtraction, based on their semantic similarity.

There are two approaches for terminal replacement. The first is straightforward: a terminal is randomly selected from the new terminal class. The second approach uses an archive strategy. Specifically, an archive is maintained to store the best individuals discovered during the GPDR process (see \ref{Archive strategy}). We measure the importance of candidate terminals based on how frequently they appear in individuals in the archive. Then, a terminal is selected randomly using roulette wheel selection, where the probability of selection increases with the frequency of the terminal's appearance. We prefer this frequency-based method over a contribution-based method \cite{mei2016feature}, as it is computationally more efficient while providing adequate accuracy. In the initialization phase, the dimension repair procedure randomly selects terminals from the new terminal class, since the archive has not yet been updated. During the breeding stage of each generation, the terminals are selected based on the frequencies in the archive.

\subsubsection{Dimensionally consistent individual generation}
\label{Algorithm framework}

The pseudocode for the dimension repair procedure is shown in Algorithm \ref{Algorithm_Dimension Repair Operator}. It takes a GP individual as input, which may consist of multiple expression trees. For each tree in the individual, if the tree is already dimensionally consistent, no further action is needed.
If the tree is dimensionally inconsistent, an MILP model is built based on the structure of the tree and its target dimension vector, as described in Section \ref{MILP Model}. This model is then solved by a general MILP solver (lines 4-7). The size and computational complexity of the model depend on the size of the expression tree. In most cases, the solver can successfully find the optimal solution. However, in cases where the model is infeasible, often due to the tree being too small or lacking the necessary structure to achieve dimensional consistency, a subtree mutation operator is applied to modify the tree's structure and make the model feasible (lines 8-9). Once a feasible solution is found, the nodes within the tree that require adjustment are identified. These nodes are replaced with new functions or terminals based on the node replacement strategy outlined in Section \ref{Node Replacement Strategy} to ensure the individual remains dimensionally consistent (lines 10-17). 

\begin{algorithm}
\caption{Dimension Repair Procedure}
\label{Algorithm_Dimension Repair Operator}
\begin{algorithmic}[1]
	\STATE \textbf{Input}: Individual $Ind$
	\STATE \textbf{Output}: Dimensionally consistent individual $Ind$
	\FOR{each $Tree \in Ind$}
        \WHILE{$Tree$ is dimensionally inconsistent}
            \STATE Let $N$ and $\mathbf{d}*$ be the node set and target dimension vector of $Tree$
            \STATE $Model.build(N,\mathbf{d}*)$
            \STATE $Model.optimize()$
            \IF{$Model$ is infeasible} 
                \STATE $Tree.SubtreeMutate()$
            \ELSE
                \FOR{each $n \in N$}
                    \IF{$x_n=1$}
                        \STATE Replace node $n$ based on the value of $y_{n,f}$ and $z_{n,t}$ with strategies described in \ref{Node Replacement Strategy}
                    \ENDIF
                \ENDFOR
            \ENDIF
        \ENDWHILE
    \ENDFOR
\end{algorithmic}
\end{algorithm}

\subsection{Archive Update}
\label{Archive strategy}

To support the dimension repair procedure in selecting terminals for replacement, an external archive is used to store the Pareto front identified up to that point based on fitness and program size. Moreover, maintaining the archive aids in controlling individual bloat \cite{wang2022multi}, thereby enhancing the interpretability of individuals and reducing the computational cost of the dimension repair procedure.
Algorithm \ref{Algorithm_Archive Update Algorithm} outlines the archive update procedure. Non-dominated sorting is applied to the population to identify non-dominated individuals. These individuals are added to the archive if they are neither dominated by nor duplicates of any other individual in the archive. 

The duplication-checking process is as follows. The output of an AR can be represented as a vector $[A_0,\dots,A_{P-1}]$, which defines an appointment schedule for $P$ patients. If the distance between the appointment schedules generated by two individuals is 0, these two individuals are considered duplicates, with the individual having the larger program size being removed.

\begin{algorithm}
\caption{Archive Update Procedure}
\label{Algorithm_Archive Update Algorithm}
\begin{algorithmic}[1]
	\STATE \textbf{Input}: Population $Pop$, $Archive$
	\STATE \textbf{Output}: Updated $Archive$
	\STATE $Pop_{nd} \gets NonDominatedSorting(Pop)$
    \FOR{each $Ind \in Pop_{nd}$}
        \STATE $dominated \gets false$
        \STATE $duplicate \gets false$
        \FOR{each $Ind' \in Archive$}
            \IF{$Ind$ is dominated by $Ind'$}
                \STATE $dominated \gets true$
                \STATE break
            \ELSIF{$Ind'$ is dominated by $Ind$}
                \STATE $Archive.remove(Ind')$
            \ENDIF
            \IF{$Ind$ is duplicate to $Ind'$}
                \IF{$Ind$ is shorter than $Ind'$}
                    \STATE $Archive.remove(Ind')$
                \ELSE
                    \STATE $duplicate \gets true$
                \ENDIF
            \ENDIF
        \ENDFOR
        \IF{$dominated=false$ and $duplicate=false$}
            \STATE $Archive.add(Ind)$
        \ENDIF
    \ENDFOR
    \RETURN $Archive$
\end{algorithmic}
\end{algorithm}

\subsection{Genetic operators}
\label{Genetic operators}

GPDR uses crossover and mutation operators to generate offspring. For mutation, the subtree mutation operator is adopted, which is performed by replacing a randomly selected subtree of an individual with a newly randomly generated subtree. For crossover, the CS crossover operator is adopted \cite{zhang2018genetic}, which is proven to be able to achieve a balance between performance and program size compared to other crossover operators used for multi-tree representations\cite{zhu2024crossover}. It randomly selects one tree for crossover and another tree for swapping. 

\section{Experimental Studies}
\label{Experimental Studies}

\subsection{Simulation Model}
\label{Simulation Model}

In our experiments, we use a single-server, single-stage queueing system to represent the simulated clinic, following the classic settings established in the literature \cite{cayirli2006designing,cayirli2012universal}. An appointment session of 3.5 hours (210 minutes) is used to represent a half-day period. We defined two levels of clinic size, the number of patients $P = 10 \text{ and } 20$, corresponding to mean service times of $M = 21.0 \text{ and } 10.5$ minutes, respectively. The coefficient of variation of service times is set at three levels, $CV = 0.4, 0.6 \text{ and } 0.8$, based on empirical values reported in the literature, which range from approximately 0.35 to 0.85 \cite{cayirli2006designing}. Service times follow a log-normal distribution as supported by empirical data \cite{klassen1996scheduling}.
The probability of no-shows $PN$ and the probability of walk-ins $PW$ are both defined at two levels, 0\% and 15\%. Patients with appointments are expected to arrive punctually. Although this assumption may seem unrealistic, previous studies have shown that unpunctuality is a less critical factor in designing AR, as most patients tend to arrive early rather than late \cite{cayirli2006designing}. The arrival of walk-in patients follows a Poisson distribution. Walk-in patients are given the lowest priority, but they will not have to wait for more than three scheduled patients to be seen \cite{cayirli2006designing}. If the clinic is empty, walk-in patients will be seen sooner. Lastly, the cost ratio $CR$ is fixed at 0.1. In total, 24 simulated clinics were modeled in our study, considering two levels of mean service times, three levels of the coefficient of variation, and two levels each for the probability of no-shows and walk-ins.

\subsection{Parameter Setting}
\label{Parameter Setting}

Table \ref{Table_Terminal set} presents the set of terminals used in this study. These terminals are derived from the problem domain. Note that $T$ and $CV$ are not included, as $T$ can be simply derived from $P \times M$, and $CV$ from $V / M$. The dimensions of $M$ and $V$ are 1, while the dimensions of the other terminals are 0. 
The function set used in this study includes Add, Sub, Mul, Div, Max, Min, Sqrt, and if-else-then. Their corresponding dimensional operations are shown in Table \ref{Table_Functions}. Note that Sq is not used, as it can be easily replaced by multiplication. The Div operator is protected and returns a large number if divided by zero.
The parameter settings for GPDR are shown in Table \ref{Table_Parameter Settings}. Most of these parameters follow the classic settings established in the literature \cite{koza1992genetic}. 

\begin{table}[!htp]\centering
\caption{Terminal set}
\setlength{\tabcolsep}{1mm}{
\begin{tabular}{llc}

\toprule
Terminals & Description & Dimension\\
\midrule
$P$ & Number of patients & 0\\
$i$ & Index of patients & 0\\
$M$ & Mean service time of patients & 1\\
$V$ & Standard deviation of patients & 1\\
$PN$ & Probability of no-shows & 0\\
$PW$ & Probability of walk-ins & 0\\
$CR$ & Cost ratio & 0\\
$\alpha$ & Random integers in the range of 0 to 2 & 0 \\
$\beta$ & Random real numbers in the range of 0 to 1 & 0 \\
\bottomrule

\end{tabular}}
\label{Table_Terminal set}
\end{table}

\begin{table}[!htp]\centering
\caption{Parameter Settings}
\setlength{\tabcolsep}{2mm}{
\begin{tabular}{ll}

\toprule
Parameter & Value\\
\midrule
Population size & 256\\
Maximum generations & 50\\
Initialization method & \makecell[l]{ramped-half-and-half \\ (minimum/maximum depth 2/6)} \\ 
Maximum depth & 8 \\ 
Selection method & tournament selection (size 7) \\
Crossover rate & 90\% \\ 
Mutation rate & 10\% \\ 
\bottomrule

\end{tabular}}
\label{Table_Parameter Settings}
\end{table}

\subsection{Comparison Design}
\label{Comparison Design}

To verify the effectiveness of the proposed GPDR, we design two sets of experiments.

\begin{enumerate}
    \item Compare with traditional manually designed AR as listed in Table \ref{Table_AR}, to assess the quality of the AR evolved by GPDR.
    \item Compare with the state-of-the-art DAGP methods, including weak-typed DAGP like Culling DAGP (CUGP) \cite{keijzer1999dimensionally}, Constrained DAGP (CDAGP) \cite{mei2017constrained}, and multi-objective DAGP (MOGP) \cite{keijzer2000genetic}, as well as strong-typed DAGP like strong-typed GP (STGP) \cite{hunt2016evolving}, and context-free grammar GP (CFGGP) \cite{da2021using}. Finally, standard GP is also included in the comparison as a benchmark for evaluating test performance and dimensional consistency. All GP variants use identical parameters and individual representations to ensure a fair comparison.
\end{enumerate}

Note that CFGGP uses the random initialization method proposed in \cite{ratle2000genetic} for controlling the derivation depth, and adopted a balanced grammar as described in \cite{da2021using}. The specific grammar is presented in online supplement. 

For weakly-typed DAGP methods and standard GP, dimensional inconsistency is evaluated during both training and testing using the dimension gaps, as proposed by \cite{mei2017constrained}. In this work, we adapt the original functions to account for the target output dimensions of expression trees, as well as the multi-tree individual representations.
If the function class $f_n$ of node $n$ is 1 or 6, and its children have inconsistent dimensions, the dimension of $n$ is set to the average dimension of its children. The dimension gap of node $n$, denoted as $DimGap(n)$, is defined as the Manhattan distance between the dimension vectors of its children. For nodes with function classes 2-5, there are no restrictions on the dimensions of their children, so their dimension gaps are always 0, as shown in  Equation (21). 
The dimension gap of expression tree $Tree$, denoted as $DimGap(Tree)$, is calculated by summing the dimension gaps of all nodes in the tree, plus the Manhattan distance between the dimension vector of its root node $n*$ and the target dimension vector $d*$, as shown in Equation (22). Since an individual can be composed of multiple trees, the dimension gap of individual $Ind$, denoted as $DimGap(Ind)$, is the sum of the dimension gaps of all the trees that make up the individual, as shown in Equation (23). This provides a measure of the dimensional inconsistency of individuals. The greater the $DimGap(Ind)$, the higher the dimensional inconsistency of individual $Ind$.

\begin{flalign}
& DimGap(n) & = &
\begin{cases}
|| \mathbf{d}_{n_1} - \mathbf{d}_{n_2}||,\\ \text{ if } f_n = 1, \mathbf{c}_n= \{ n_1, n_2 \} \\
|| \mathbf{d}_{n_2} - \mathbf{d}_{n_3}||,\\ \text{ if } f_n = 6, \mathbf{c}_n= \{ n_1, n_2, n_3 \} \\
0, \text{ otherwise}
\end{cases} \\
& DimGap(Tree) & = & \sum_{n \in Tree} DimGap(n) + || \mathbf{d}_{n*} - \mathbf{d}*|| \\
& DimGap(Ind)  & = & \sum_{Tree \in Ind} DimGap(Tree)
\end{flalign}

Each algorithm is independently run 30 times for each simulated clinic. Different random seeds are used in every training generation and experiment. The test performance is measured by the average performance and dimension gap of the 30 AR generated during these runs. All DAGP methods were implemented using Evolutionary Computation in Java, as described in \cite{luke2017ecj}. The experimental tests were conducted on a system equipped with an Intel Core™ i7-10700 CPU running at 2.90GHz, and 16GB of RAM. The solver used in this study is Gurobi 12.01.

\subsection{Results and Discussions}
\label{Results and Discussions}

In this section, we conduct two comparisons to assess the effectiveness of the newly proposed GPDR: one against traditional manually designed AR and another against other DAGP methods.

\subsubsection{Comparison with manually designed AR}
\label{Comparison with manual designed AR}

Table \ref{Table_Comparison with manual designed AR} shows the performance comparison between traditional manually designed and evolved AR. In each simulated clinic, GPDR was independently run 30 times, and the mean result (with standard deviations in parentheses) was compared with the manually designed AR. Each rule was simulated another 15,000 replications to guarantee that the results were within a 1\% error margin at a 95\% confidence level. The best test performance for each simulated clinic is highlighted in bold. We conducted statistical significance testing using a one-sample t-test with a significance level of 0.05. Here, $+$ indicates that GPDR performs significantly better than the compared rule, $=$ means no significant difference, and $-$ denotes that GPDR performs significantly worse than the compared rule. From Table \ref{Table_Comparison with manual designed AR}, it is evident that the evolved AR consistently outperforms the manually designed AR for all simulated clinics, demonstrating that GPDR can generate better efficient frontiers.

\begin{table*}\centering
\caption{Test performance of manually designed and evolved AR}
\setlength{\tabcolsep}{2.5mm}{
\begin{tabular}{cccccccc}

\toprule
Clinic & IBFI  & 2BEG  & MBFI  & OFFSET & DOME  & RULE7 & GPDR \\
\midrule
\textless M=21.0,CV=0.40,PN=0.00,PW=0.00\textgreater & 13.1473(+)  & 22.7486(+)  & 21.4672(+)  & 12.2431(+)  & 12.7261(+)  & 16.8433(+) & \textbf{12.0598(0.1803)} \\
\textless M=21.0,CV=0.40,PN=0.00,PW=0.15\textgreater & 29.6358(+)  & 36.8856(+)  & 36.7445(+)  & 28.0179(+)  & 28.7210(+)  & 31.8029(+) & \textbf{27.7458(0.2330)} \\
\textless M=21.0,CV=0.40,PN=0.15,PW=0.00\textgreater & 12.2495(+)  & 17.2875(+)  & 19.2849(+)  & 12.7861(+)  & 12.6909(+)  & 14.4997(+) & \textbf{12.0786(0.0323)} \\
\textless M=21.0,CV=0.40,PN=0.15,PW=0.15\textgreater & 28.0619(+)  & 29.5890(+)  & 33.4897(+)  & 28.3758(+)  & 28.4759(+)  & 28.6904(+) & \textbf{27.1289(0.0509)} \\
\textless M=21.0,CV=0.60,PN=0.00,PW=0.00\textgreater & 19.1538(+)  & 26.2911(+)  & 26.5679(+)  & 18.0563(+)  & 18.7103(+)  & 20.3627(+) & \textbf{17.8036(0.2466)} \\
\textless M=21.0,CV=0.60,PN=0.00,PW=0.15\textgreater & 35.0325(+)  & 39.8933(+)  & 41.3343(+)  & 33.1462(=)  & 34.0662(+)  & 34.9180(+) & \textbf{33.1269(0.4317)} \\
\textless M=21.0,CV=0.60,PN=0.15,PW=0.00\textgreater & 16.9983(+)  & 20.9115(+)  & 23.2240(+)  & 17.7745(+)  & 17.6731(+)  & 18.4870(+) & \textbf{16.6983(0.0648)} \\
\textless M=21.0,CV=0.60,PN=0.15,PW=0.15\textgreater & 32.5623(+)  & 33.3586(+)  & 37.2810(+)  & 33.1281(+)  & 33.2621(+)  & 33.1490(+) & \textbf{31.3892(0.0683)} \\
\textless M=21.0,CV=0.80,PN=0.00,PW=0.00\textgreater & 24.5980(+)  & 30.1703(+)  & 31.2582(+)  & 23.4816(+)  & 24.2906(+)  & 24.8229(+) & \textbf{23.0391(0.2501)} \\
\textless M=21.0,CV=0.80,PN=0.00,PW=0.15\textgreater & 40.0632(+)  & 43.4678(+)  & 45.7052(+)  & 38.1565(=)  & 39.2652(+)  & 39.1283(+) & \textbf{38.0848(0.3822)} \\
\textless M=21.0,CV=0.80,PN=0.15,PW=0.00\textgreater & 21.6863(+)  & 24.8108(+)  & 27.2603(+)  & 22.7693(+)  & 22.6756(+)  & 23.2970(+) & \textbf{21.1649(0.1117)} \\
\textless M=21.0,CV=0.80,PN=0.15,PW=0.15\textgreater & 37.0495(+)  & 37.3219(+)  & 41.2120(+)  & 38.0191(+)  & 38.1683(+)  & 38.1703(+) & \textbf{35.6691(0.0781)} \\
\textless M=10.5,CV=0.40,PN=0.00,PW=0.00\textgreater & 9.0091(+)  & 13.0627(+)  & 13.2318(+)  & 7.5418(+)  & 9.6934(+)  & 8.8472(+) & \textbf{7.3621(0.2732)} \\
\textless M=10.5,CV=0.40,PN=0.00,PW=0.15\textgreater & 23.7402(+)  & 27.2990(+)  & 27.3600(+)  & 20.6513(+)  & 24.3200(+)  & 21.8317(+) & \textbf{19.3406(0.2858)} \\
\textless M=10.5,CV=0.40,PN=0.15,PW=0.00\textgreater & 6.7667(+)  & 8.3368(+)  & 10.3017(+)  & 7.0043(+)  & 7.1954(+)  & 7.5482(+) & \textbf{6.5633(0.0388)} \\
\textless M=10.5,CV=0.40,PN=0.15,PW=0.15\textgreater & 18.6936(+)  & 19.3621(+)  & 21.6000(+)  & 17.5780(=)  & 19.2999(+)  & 18.0191(+) & \textbf{17.5473(0.1112)} \\
\textless M=10.5,CV=0.60,PN=0.00,PW=0.00\textgreater & 13.2060(+)  & 16.1769(+)  & 17.0009(+)  & 11.1332(=)  & 14.2844(+)  & 12.3891(+) & \textbf{11.0922(0.1698)} \\
\textless M=10.5,CV=0.60,PN=0.00,PW=0.15\textgreater & 27.1588(+)  & 29.5796(+)  & 30.4072(+)  & 23.1983(+)  & 28.1302(+)  & 24.2218(+) & \textbf{22.7243(0.2337)} \\
\textless M=10.5,CV=0.60,PN=0.15,PW=0.00\textgreater & 9.8103(+)  & 11.0673(+)  & 12.9583(+)  & 9.9202(+)  & 10.5905(+)  & 10.7869(+) & \textbf{9.3855(0.0831)} \\
\textless M=10.5,CV=0.60,PN=0.15,PW=0.15\textgreater & 21.6366(+)  & 22.0805(+)  & 24.2455(+)  & 20.1694(+)  & 22.6219(+)  & 21.1037(+) & \textbf{20.3597(0.2152)} \\
\textless M=10.5,CV=0.80,PN=0.00,PW=0.00\textgreater & 17.0779(+)  & 19.3996(+)  & 20.5160(+)  & 14.5857(+)  & 18.6032(+)  & 15.2600(+) & \textbf{14.7771(0.1758)} \\
\textless M=10.5,CV=0.80,PN=0.00,PW=0.15\textgreater & 30.4853(+)  & 32.2145(+)  & 33.4390(+)  & 25.9640(+)  & 31.9229(+)  & 26.4166(+) & \textbf{26.2221(0.3146)} \\
\textless M=10.5,CV=0.80,PN=0.15,PW=0.00\textgreater & 12.9671(+)  & 13.9992(+)  & 15.8024(+)  & 12.9563(+)  & 14.1452(+)  & 13.7264(+) & \textbf{12.2799(0.0812)} \\
\textless M=10.5,CV=0.80,PN=0.15,PW=0.15\textgreater & 24.7033(+)  & 25.0002(+)  & 27.0746(+)  & 23.0097(+)  & 26.1190(+)  & 23.8726(+) & \textbf{23.2157(0.2992)} \\
\midrule
    Avg   & 21.8956  & 25.0131  & 26.6153  & 20.8194  & 22.4021  & 22.0081  & \textbf{20.2858(0.1838)} \\
\bottomrule

\end{tabular}}
\label{Table_Comparison with manual designed AR}
\end{table*}

\subsubsection{Comparison with other DAGP methods}
\label{Comparison with other DAGP methods}

Table \ref{Table_Comparison with other DAGP methods} shows the performance comparison between the proposed GPDR, other DAGP methods and standard GP, focusing on two key metrics: test performance and dimension gap. In both cases, smaller values indicate better performance. The table reports the mean test performance (with standard deviations in parentheses) over 30 runs for each method in 24 simulated clinics. For weakly-typed DAGP methods, their mean dimension gaps are reported below the test performance. We also show dimension gaps of 0 below strongly-typed DAGP methods for comparison. We conducted statistical significance tests using the Wilcoxon signed rank test at a 0.05 significance level. As mentioned before, the symbols next to each algorithm indicate whether GPDR performs significantly better than, worse than, or comparable to the compared algorithm.

From Table \ref{Table_Comparison with other DAGP methods}, we can draw the following observations. First, GPDR shows no significant difference in most cases and is only slightly outperformed by standard GP in a few clinics, with minimal average differences in test performance. Given that the awareness of dimensions reduces the search space for GP, limiting the potential for finding better solutions, the performance of GPDR is impressive. For weakly-typed DAGP methods, our GPDR consistently achieves performance that is either comparable to or significantly better than CUGP, CDAGP, and MOGP in all simulated clinics. For strongly-typed DAGP methods, our GPDR consistently outperforms STGP and CFGGP in all simulated clinics, demonstrating significantly better results. 
Secondly, for weakly-typed DAGP methods, CUGP achieves the best test performance but suffers from the highest dimension gap. MOGP maintains a low level of dimension gap, but its performance is relatively poor. CDAGP lies in the middle of other weakly typed DAGP methods. For strongly-typed DAGP methods, STGP and CFGGP can maintain the dimension consistency of individuals but have poor test performance. 
In summary, GPDR achieves the best trade-off between test performance and dimension gap. GPDR not only maintains dimension consistency as a strongly typed DAGP, but also exhibits a search capability on par with that of weakly typed DAGP methods and standard GP. 

\begin{table*}\centering
\caption{Mean (Std.) Test performance, significance test and dimension gap of DAGP methods}
\setlength{\tabcolsep}{1mm}{
\resizebox{\textwidth}{!}{
\begin{tabular}{cccccccc}

\toprule
Clinic & Standard GP & CUGP\cite{keijzer1999dimensionally}  & CDAGP\cite{mei2017constrained} & MOGP\cite{keijzer2000genetic}  & STGP\cite{hunt2016evolving}  & CFGGP\cite{da2021using} & GPDR \\
\midrule
\textless M=21.0,CV=0.40,PN=0.00,PW=0.00\textgreater & \makecell{12.0507(0.0856)(=)\\2.3898} & \makecell{12.0822(0.1004)(+)\\2.4167} & \makecell{12.1280(0.1163)(+)\\4.7570} & \makecell{12.3150(0.4396)(+)\\0.5667} & \makecell{12.4148(0.3063)(+)\\0} & \makecell{12.4617(0.2887)(+)\\0} & \makecell{12.0598(0.1803)\\0} \\
\textless M=21.0,CV=0.40,PN=0.00,PW=0.15\textgreater & \makecell{27.8096(0.2966)(=)\\2.6250} & \makecell{27.7366(0.2310)(+)\\1.9680} & \makecell{27.8073(0.3115)(=)\\2.0120} & \makecell{27.9595(0.6664)(=)\\0.4667} & \makecell{28.4812(0.5053)(+)\\0} & \makecell{28.9471(0.3769)(+)\\0} & \makecell{27.7458(0.2330)\\0} \\
\textless M=21.0,CV=0.40,PN=0.15,PW=0.00\textgreater & \makecell{12.0610(0.0355)(=)\\2.5680} & \makecell{12.0962(0.0369)(+)\\2.5482} & \makecell{12.1143(0.0612)(+)\\1.2625} & \makecell{12.1180(0.1041)(+)\\0.0667} & \makecell{12.1287(0.0262)(+)\\0} & \makecell{12.1667(0.0406)(+)\\0} & \makecell{12.0786(0.0323)\\0} \\
\textless M=21.0,CV=0.40,PN=0.15,PW=0.15\textgreater & \makecell{27.1093(0.0446)(-)\\3.6219} & \makecell{27.2269(0.0537)(+)\\4.4172} & \makecell{27.2255(0.0551)(+)\\1.9906} & \makecell{27.1967(0.1770)(=)\\0.1000} & \makecell{27.2543(0.0836)(+)\\0} & \makecell{27.6750(0.2571)(+)\\0} & \makecell{27.1289(0.0509)\\0} \\
\textless M=21.0,CV=0.60,PN=0.00,PW=0.00\textgreater & \makecell{17.7447(0.1528)(=)\\3.2844} & \makecell{17.7708(0.1547)(=)\\3.5612} & \makecell{17.8125(0.1807)(=)\\1.5292} & \makecell{18.1267(0.5296)(+)\\0.4667} & \makecell{18.1268(0.3414)(+)\\0} & \makecell{18.4654(0.3030)(+)\\0} & \makecell{17.8036(0.2466)\\0} \\
\textless M=21.0,CV=0.60,PN=0.00,PW=0.15\textgreater & \makecell{33.0157(0.3039)(=)\\3.2393} & \makecell{33.0671(0.3419)(=)\\2.4031} & \makecell{33.0740(0.3614)(=)\\3.4424} & \makecell{33.5675(0.7362)(+)\\0.3333} & \makecell{33.8678(0.4452)(+)\\0} & \makecell{34.3196(0.3276)(+)\\0} & \makecell{33.1269(0.4317)\\0} \\
\textless M=21.0,CV=0.60,PN=0.15,PW=0.00\textgreater & \makecell{16.5990(0.0918)(-)\\2.7185} & \makecell{16.6580(0.0899)(=)\\2.6607} & \makecell{16.6676(0.1064)(=)\\4.3883} & \makecell{16.7241(0.0448)(=)\\0.0667} & \makecell{16.7821(0.0623)(+)\\0} & \makecell{16.8495(0.0754)(+)\\0} & \makecell{16.6983(0.0648)\\0} \\
\textless M=21.0,CV=0.60,PN=0.15,PW=0.15\textgreater & \makecell{31.3984(0.0642)(=)\\3.6956} & \makecell{31.4773(0.0625)(+)\\4.0552} & \makecell{31.5365(0.0689)(+)\\1.8982} & \makecell{31.4221(0.0607)(+)\\0.2000} & \makecell{31.5826(0.1573)(+)\\0} & \makecell{32.1682(0.3581)(+)\\0} & \makecell{31.3892(0.0683)\\0} \\
\textless M=21.0,CV=0.80,PN=0.00,PW=0.00\textgreater & \makecell{23.0585(0.2234)(=)\\2.8563} & \makecell{23.0250(0.2206)(+)\\3.0646} & \makecell{23.2059(0.2262)(+)\\1.2471} & \makecell{23.4598(0.5709)(+)\\0.4333} & \makecell{23.6241(0.4242)(+)\\0} & \makecell{24.0034(0.2298)(+)\\0} & \makecell{23.0391(0.2501)\\0} \\
\textless M=21.0,CV=0.80,PN=0.00,PW=0.15\textgreater & \makecell{37.9690(0.3840)(=)\\3.0143} & \makecell{38.0335(0.4201)(=)\\2.7797} & \makecell{38.2295(0.3843)(=)\\1.5544} & \makecell{38.6600(0.6839)(+)\\0.3000} & \makecell{38.9874(0.4780)(+)\\0} & \makecell{39.4092(0.4137)(+)\\0} & \makecell{38.0848(0.3822)\\0} \\
\textless M=21.0,CV=0.80,PN=0.15,PW=0.00\textgreater & \makecell{21.0162(0.0856)(-)\\2.1711} & \makecell{21.0979(0.1171)(=)\\2.0245} & \makecell{21.1806(0.1233)(=)\\1.2185} & \makecell{21.2317(0.1021)(+)\\0.1333} & \makecell{21.3214(0.1079)(+)\\0} & \makecell{21.4561(0.0746)(+)\\0} & \makecell{21.1649(0.1117)\\0} \\
\textless M=21.0,CV=0.80,PN=0.15,PW=0.15\textgreater & \makecell{35.6761(0.0799)(=)\\3.6307} & \makecell{35.7918(0.0759)(+)\\3.6055} & \makecell{35.8011(0.0775)(+)\\2.3375} & \makecell{35.7342(0.1988)(=)\\0.2667} & \makecell{35.9020(0.1505)(+)\\0} & \makecell{36.6648(0.3126)(+)\\0} & \makecell{35.6691(0.0781)\\0} \\
\textless M=10.5,CV=0.40,PN=0.00,PW=0.00\textgreater & \makecell{7.3461(0.0722)(=)\\2.1891} & \makecell{7.3572(0.0723)(+)\\2.2331} & \makecell{7.4128(0.0631)(+)\\2.9794} & \makecell{7.6691(0.5749)(+)\\0.6000} & \makecell{7.6862(0.4512)(+)\\0} & \makecell{7.6512(0.4642)(+)\\0} & \makecell{7.3621(0.2732)\\0} \\
\textless M=10.5,CV=0.40,PN=0.00,PW=0.15\textgreater & \makecell{19.6488(0.5594)(+)\\3.2466} & \makecell{19.6478(0.5747)(=)\\3.0661} & \makecell{19.6785(0.5576)(+)\\2.6161} & \makecell{19.7586(0.4506)(+)\\0.3333} & \makecell{19.9257(0.5382)(+)\\0} & \makecell{21.1764(1.2287)(+)\\0} & \makecell{19.3406(0.2858)\\0} \\
\textless M=10.5,CV=0.40,PN=0.15,PW=0.00\textgreater & \makecell{6.5402(0.0279)(-)\\3.1188} & \makecell{6.5510(0.0363)(=)\\2.8513} & \makecell{6.5615(0.0284)(=)\\3.1445} & \makecell{6.6168(0.0512)(+)\\0.1667} & \makecell{6.6408(0.0304)(+)\\0} & \makecell{6.6720(0.0235)(+)\\0} & \makecell{6.5633(0.0388)\\0} \\
\textless M=10.5,CV=0.40,PN=0.15,PW=0.15\textgreater & \makecell{17.4872(0.0727)(-)\\4.0831} & \makecell{17.5170(0.0827)(=)\\3.3661} & \makecell{17.5344(0.1068)(=)\\3.0281} & \makecell{17.7738(0.3384)(+)\\0.3667} & \makecell{17.9995(0.2795)(+)\\0} & \makecell{18.2810(0.1030)(+)\\0} & \makecell{17.5473(0.1112)\\0} \\
\textless M=10.5,CV=0.60,PN=0.00,PW=0.00\textgreater & \makecell{11.1187(0.1451)(=)\\2.1120} & \makecell{11.1769(0.1439)(=)\\2.7404} & \makecell{11.1665(0.1545)(=)\\2.8469} & \makecell{11.4828(0.6097)(+)\\0.4667} & \makecell{11.4709(0.5077)(+)\\0} & \makecell{11.6058(0.5593)(+)\\0} & \makecell{11.0922(0.1698)\\0} \\
\textless M=10.5,CV=0.60,PN=0.00,PW=0.15\textgreater & \makecell{22.9300(0.4905)(=)\\3.4031} & \makecell{22.8884(0.5192)(=)\\5.1586} & \makecell{23.0832(0.5929)(+)\\3.2232} & \makecell{23.5721(1.1752)(+)\\0.2333} & \makecell{23.4260(0.8856)(+)\\0} & \makecell{24.3874(0.7387)(+)\\0} & \makecell{22.7243(0.2337)\\0} \\
\textless M=10.5,CV=0.60,PN=0.15,PW=0.00\textgreater & \makecell{9.3399(0.0599)(-)\\4.4154} & \makecell{9.3832(0.0808)(=)\\3.7266} & \makecell{9.4202(0.0831)(+)\\1.3508} & \makecell{9.5114(0.1104)(+)\\0.2333} & \makecell{9.5361(0.1005)(+)\\0} & \makecell{9.6217(0.0680)(+)\\0} & \makecell{9.3855(0.0831)\\0} \\
\textless M=10.5,CV=0.60,PN=0.15,PW=0.15\textgreater & \makecell{20.2481(0.1389)(-)\\3.7419} & \makecell{20.2498(0.0644)(=)\\3.3716} & \makecell{20.3539(0.2022)(=)\\2.8206} & \makecell{20.4854(0.2821)(=)\\0.3000} & \makecell{20.8519(0.3590)(+)\\0} & \makecell{21.1222(0.2729)(+)\\0} & \makecell{20.3597(0.2152)\\0} \\
\textless M=10.5,CV=0.80,PN=0.00,PW=0.00\textgreater & \makecell{14.7640(0.2147)(=)\\2.0307} & \makecell{14.8143(0.2259)(+)\\2.2487} & \makecell{14.8736(0.2201)(+)\\1.5695} & \makecell{15.2306(0.6790)(+)\\0.3333} & \makecell{15.1893(0.4379)(+)\\0} & \makecell{15.3152(0.5354)(+)\\0} & \makecell{14.7771(0.1758)\\0} \\
\textless M=10.5,CV=0.80,PN=0.00,PW=0.15\textgreater & \makecell{26.2468(0.5463)(=)\\4.3570} & \makecell{26.1932(0.3730)(=)\\3.7651} & \makecell{26.5102(0.6750)(=)\\2.3510} & \makecell{26.9349(0.8974)(+)\\0.3000} & \makecell{26.7249(0.7633)(+)\\0} & \makecell{28.2496(0.9933)(+)\\0} & \makecell{26.2221(0.3146)\\0} \\
\textless M=10.5,CV=0.80,PN=0.15,PW=0.00\textgreater & \makecell{12.2358(0.0982)(=)\\3.7039} & \makecell{12.2392(0.0801)(=)\\3.4477} & \makecell{12.2756(0.0905)(=)\\3.5815} & \makecell{12.4620(0.1673)(+)\\0.2333} & \makecell{12.5168(0.1506)(+)\\0} & \makecell{12.7006(0.0895)(+)\\0} & \makecell{12.2799(0.0812)\\0} \\
\textless M=10.5,CV=0.80,PN=0.15,PW=0.15\textgreater & \makecell{23.1309(0.1255)(=)\\4.4818} & \makecell{23.1460(0.1147)(=)\\3.8951} & \makecell{23.2430(0.1939)(=)\\2.8372} & \makecell{23.4760(0.4295)(+)\\0.3667} & \makecell{23.6873(0.4052)(+)\\0} & \makecell{24.0967(0.2698)(+)\\0} & \makecell{23.2157(0.2992)\\0} \\
\midrule
Avg   & \makecell{20.2727(0.1833)\\3.1958} & \makecell{20.3011(0.1780)\\3.1406} & \makecell{20.3707(0.2100)\\2.4994} & \makecell{20.5620(0.4200)\\0.3056} & \makecell{20.6720(0.3332)\\0} & \makecell{21.0611(0.3502)\\0} & \makecell{20.2858(0.1838)\\0} \\

\bottomrule

\end{tabular}}}
\label{Table_Comparison with other DAGP methods}
\end{table*}

\section{Further Analysis}
\label{Further Analysis}

\subsection{Program Size}
\label{Program Size}

The program size is defined as the number of nodes in an individual's expression trees. A smaller number of nodes reduces the efforts needed for a human user to simulate AR’s behavior, enhancing its interpretability \cite{mei2022explainable}. 
Table \ref{Table_Program size} shows the mean program size of the proposed GPDR, other DAGP methods and standard GP. As mentioned before, the symbols next to each algorithm indicate whether GPDR performs significantly better than, worse than, or comparable to the compared algorithm. MOGP performs the best in terms of program size, while CFGGP tends to produce very deep and large trees. This is because the proportion of terminal derivations in CFG trees is lower than that in standard GP expression trees. GPDR achieves better program size than CUGP and STGP, but falls slightly short compared to standard GP and CDAGP. This indicates that the improved performance of GPDR comes at the expense of program size. The dimension repair procedure in GPDR leads to an increase in the program size of individuals.

\begin{table*}\centering
\caption{Mean (Std.) Program Size and significance test of DAGP methods}
\setlength{\tabcolsep}{1mm}{
\resizebox{\textwidth}{!}{
\begin{tabular}{cccccccc}

\toprule
Clinic & Standard GP & CUGP\cite{keijzer1999dimensionally}  & CDAGP\cite{mei2017constrained} & MOGP\cite{keijzer2000genetic}  & STGP\cite{hunt2016evolving}  & CFGGP\cite{da2021using} & GPDR \\
\midrule
\textless M=21.0,CV=0.40,PN=0.00,PW=0.00\textgreater & 25.93(25.81)(=) & 27.97(24.40)(+) & 34.97(55.93)(=) & 11.80(12.95)(-) & 56.77(25.32)(+) & 109.90(36.16)(+) & 35.53(18.88) \\
\textless M=21.0,CV=0.40,PN=0.00,PW=0.15\textgreater & 29.30(19.22)(=) & 32.33(17.70)(+) & 34.13(18.11)(=) & 20.87(12.68)(-) & 54.13(27.94)(+) & 107.27(26.82)(+) & 31.63(14.77) \\
\textless M=21.0,CV=0.40,PN=0.15,PW=0.00\textgreater & 24.97(16.25)(=) & 26.20(16.88)(+) & 21.67(22.41)(=) & 10.43(8.19)(-) & 43.17(32.11)(+) & 87.00(44.38)(+) & 24.47(14.79) \\
\textless M=21.0,CV=0.40,PN=0.15,PW=0.15\textgreater & 32.50(27.86)(=) & 32.33(21.23)(+) & 25.83(16.85)(=) & 13.60(11.64)(-) & 37.63(30.15)(=) & 104.27(45.08)(+) & 27.10(20.57) \\
\textless M=21.0,CV=0.60,PN=0.00,PW=0.00\textgreater & 33.07(26.33)(=) & 47.93(30.20)(+) & 33.17(25.17)(=) & 11.27(9.09)(-) & 50.00(34.13)(+) & 110.67(38.32)(+) & 32.53(22.29) \\
\textless M=21.0,CV=0.60,PN=0.00,PW=0.15\textgreater & 41.67(28.81)(=) & 29.70(17.31)(=) & 41.70(33.14)(=) & 17.90(12.33)(-) & 58.17(38.42)(=) & 109.67(24.73)(+) & 46.73(33.59) \\
\textless M=21.0,CV=0.60,PN=0.15,PW=0.00\textgreater & 30.23(26.64)(=) & 27.43(14.19)(+) & 35.37(52.13)(=) & 8.63(5.92)(-) & 44.27(29.77)(=) & 94.70(37.28)(+) & 33.60(21.50) \\
\textless M=21.0,CV=0.60,PN=0.15,PW=0.15\textgreater & 34.47(26.39)(+) & 39.67(24.90)(+) & 27.10(17.69)(+) & 14.97(8.76)(=) & 22.63(20.19)(=) & 102.97(43.45)(+) & 19.13(20.01) \\
\textless M=21.0,CV=0.80,PN=0.00,PW=0.00\textgreater & 32.93(26.26)(=) & 37.23(32.48)(+) & 25.70(23.05)(=) & 13.13(10.98)(-) & 39.37(28.14)(=) & 114.37(17.07)(+) & 38.40(26.89) \\
\textless M=21.0,CV=0.80,PN=0.00,PW=0.15\textgreater & 33.43(21.96)(=) & 31.90(21.20)(+) & 34.13(31.72)(=) & 14.43(8.96)(-) & 50.87(42.72)(=) & 112.50(30.24)(+) & 41.07(24.06) \\
\textless M=21.0,CV=0.80,PN=0.15,PW=0.00\textgreater & 24.10(20.81)(=) & 30.20(20.24)(+) & 20.97(17.09)(=) & 8.93(7.19)(-) & 35.83(25.48)(=) & 92.03(35.08)(+) & 29.33(29.41) \\
\textless M=21.0,CV=0.80,PN=0.15,PW=0.15\textgreater & 33.10(25.51)(=) & 35.70(24.21)(+) & 27.23(26.71)(=) & 15.43(10.52)(-) & 35.37(24.31)(=) & 98.13(38.73)(+) & 30.63(25.58) \\
\textless M=10.5,CV=0.40,PN=0.00,PW=0.00\textgreater & 23.37(20.42)(-) & 25.80(15.75)(=) & 23.73(35.62)(-) & 8.70(6.74)(-) & 58.00(39.65)(=) & 98.40(38.67)(+) & 42.40(23.35) \\
\textless M=10.5,CV=0.40,PN=0.00,PW=0.15\textgreater & 30.53(28.87)(=) & 35.37(28.41)(+) & 28.57(24.65)(=) & 18.60(12.23)(-) & 58.13(37.37)(+) & 108.13(52.15)(+) & 36.67(23.92) \\
\textless M=10.5,CV=0.40,PN=0.15,PW=0.00\textgreater & 33.10(19.08)(=) & 32.73(25.32)(+) & 30.67(34.94)(=) & 10.80(7.73)(-) & 38.67(28.24)(=) & 99.57(49.17)(+) & 31.03(14.93) \\
\textless M=10.5,CV=0.40,PN=0.15,PW=0.15\textgreater & 33.27(25.29)(=) & 34.70(24.83)(+) & 32.63(41.40)(=) & 20.60(12.75)(-) & 53.30(32.98)(=) & 100.73(31.11)(+) & 37.70(26.43) \\
\textless M=10.5,CV=0.60,PN=0.00,PW=0.00\textgreater & 24.90(22.73)(=) & 27.03(25.76)(+) & 23.90(26.79)(=) & 8.87(6.76)(-) & 54.83(32.04)(+) & 116.53(28.81)(+) & 34.40(20.08) \\
\textless M=10.5,CV=0.60,PN=0.00,PW=0.15\textgreater & 39.60(30.09)(=) & 47.30(29.62)(=) & 39.50(34.78)(=) & 13.97(8.50)(-) & 63.03(32.91)(+) & 112.03(37.84)(+) & 46.27(25.65) \\
\textless M=10.5,CV=0.60,PN=0.15,PW=0.00\textgreater & 34.33(22.39)(=) & 36.67(17.93)(+) & 21.60(14.26)(-) & 15.57(13.18)(-) & 47.03(30.47)(+) & 92.00(47.46)(+) & 30.57(15.66) \\
\textless M=10.5,CV=0.60,PN=0.15,PW=0.15\textgreater & 33.83(23.14)(=) & 38.20(20.84)(=) & 25.37(22.81)(-) & 20.00(12.69)(-) & 46.33(35.65)(=) & 111.17(35.03)(+) & 39.17(27.30) \\
\textless M=10.5,CV=0.80,PN=0.00,PW=0.00\textgreater & 22.40(21.47)(-) & 27.37(19.34)(+) & 21.03(19.41)(-) & 14.90(11.30)(-) & 53.93(37.15)(+) & 119.43(29.77)(+) & 33.50(17.97) \\
\textless M=10.5,CV=0.80,PN=0.00,PW=0.15\textgreater & 46.23(30.24)(=) & 44.27(25.94)(+) & 31.63(34.31)(=) & 20.33(20.11)(-) & 75.63(25.79)(+) & 107.13(32.24)(+) & 36.87(23.78) \\
\textless M=10.5,CV=0.80,PN=0.15,PW=0.00\textgreater & 31.93(24.15)(=) & 39.03(22.20)(+) & 36.17(36.24)(=) & 14.23(10.38)(-) & 48.20(37.97)(+) & 96.97(41.21)(+) & 28.17(14.98) \\
\textless M=10.5,CV=0.80,PN=0.15,PW=0.15\textgreater & 41.13(24.65)(=) & 39.90(31.68)(+) & 30.50(18.85)(=) & 17.43(12.71)(-) & 62.00(40.64)(+) & 110.60(40.72)(+) & 30.73(18.13) \\
\midrule
Avg   & 32.10(24.35) & 34.46(23.02) & 29.47(28.50) & 14.39(10.60) & 49.47(32.06) & 104.84(36.73) & 34.07(21.86) \\
\bottomrule
    
\end{tabular}}}
\label{Table_Program size}
\end{table*}

\subsection{Running Time}
\label{Running Time}

GPDR requires applying the dimension repair procedure to all individuals in each generation, which involves solving multiple MILP models. Therefore, we conducted a runtime analysis to assess the computational cost in training. Table \ref{Table_running time} presents the mean runtime (in seconds) for both standard GP and GPDR using 8 threads. On average, GPDR takes 10 seconds longer than standard GP, which accounts for the time spent on dimension repair. These results demonstrate that the proposed GPDR remains computationally efficient despite the added complexity.

\begin{table}[!htp]
\centering
\caption{Mean runtime (in seconds) of standard GP and GPDR}
\begin{tabular}{ccc}

\toprule
Clinic & Standard GP & GPDR \\
\midrule
\textless M=21.0,CV=0.40,PN=0.00,PW=0.00\textgreater & 32.0549  & 39.4361  \\
\textless M=21.0,CV=0.40,PN=0.00,PW=0.15\textgreater & 39.6369  & 48.4130  \\
\textless M=21.0,CV=0.40,PN=0.15,PW=0.00\textgreater & 31.0018  & 36.5804  \\
\textless M=21.0,CV=0.40,PN=0.15,PW=0.15\textgreater & 36.3241  & 42.8103  \\
\textless M=21.0,CV=0.60,PN=0.00,PW=0.00\textgreater & 35.8059  & 41.0839  \\
\textless M=21.0,CV=0.60,PN=0.00,PW=0.15\textgreater & 40.4532  & 47.7710  \\
\textless M=21.0,CV=0.60,PN=0.15,PW=0.00\textgreater & 31.0124  & 37.3955  \\
\textless M=21.0,CV=0.60,PN=0.15,PW=0.15\textgreater & 36.5554  & 42.7169  \\
\textless M=21.0,CV=0.80,PN=0.00,PW=0.00\textgreater & 34.9716  & 40.9829  \\
\textless M=21.0,CV=0.80,PN=0.00,PW=0.15\textgreater & 40.0133  & 47.2769  \\
\textless M=21.0,CV=0.80,PN=0.15,PW=0.00\textgreater & 31.3912  & 38.1833  \\
\textless M=21.0,CV=0.80,PN=0.15,PW=0.15\textgreater & 37.6199  & 44.0846  \\
\textless M=10.5,CV=0.40,PN=0.00,PW=0.00\textgreater & 62.5749  & 70.3793  \\
\textless M=10.5,CV=0.40,PN=0.00,PW=0.15\textgreater & 76.6036  & 104.4276 \\
\textless M=10.5,CV=0.40,PN=0.15,PW=0.00\textgreater & 72.5869  & 76.9220  \\
\textless M=10.5,CV=0.40,PN=0.15,PW=0.15\textgreater & 68.2984  & 82.7984  \\
\textless M=10.5,CV=0.60,PN=0.00,PW=0.00\textgreater & 64.7409  & 75.1846  \\
\textless M=10.5,CV=0.60,PN=0.00,PW=0.15\textgreater & 74.6969  & 85.1748  \\
\textless M=10.5,CV=0.60,PN=0.15,PW=0.00\textgreater & 54.7803  & 65.6362  \\
\textless M=10.5,CV=0.60,PN=0.15,PW=0.15\textgreater & 66.6318  & 78.3181  \\
\textless M=10.5,CV=0.80,PN=0.00,PW=0.00\textgreater & 64.2053  & 74.6448  \\
\textless M=10.5,CV=0.80,PN=0.00,PW=0.15\textgreater & 74.1029  & 85.0275  \\
\textless M=10.5,CV=0.80,PN=0.15,PW=0.00\textgreater & 55.2550  & 69.7782  \\
\textless M=10.5,CV=0.80,PN=0.15,PW=0.15\textgreater & 66.4251  & 91.7711  \\
\midrule
Avg   & 51.1559  & 61.1166  \\
\bottomrule
\end{tabular}
\label{Table_running time}
\end{table}

\subsection{Semantic Analysis}
\label{Semantic Analysis}

\begin{figure*}[htbp]
\includegraphics[width=\textwidth]{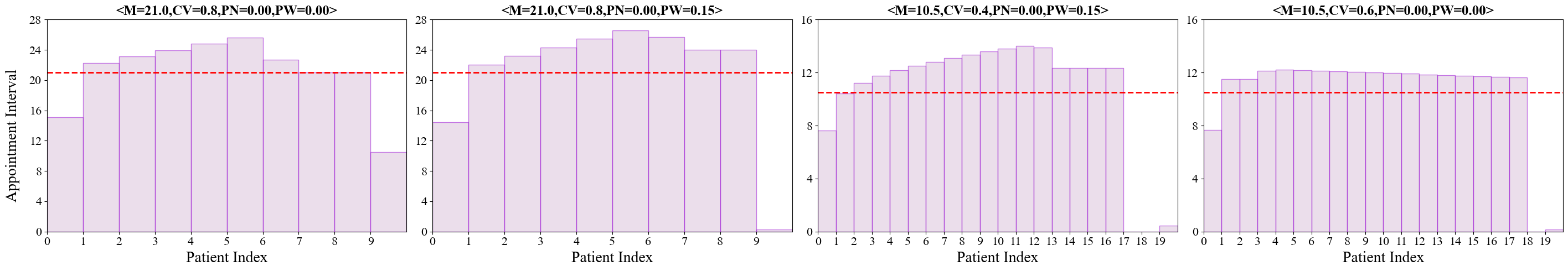}
\caption{``Dome" pattern in appointment intervals for selected simulated clinics.}
\label{Fig_plan}
\end{figure*}

To gain deeper insights into the behavior of the evolved ARs, we extracted the best AR evolved by GPDR for each simulated clinic. Figure \ref{Fig_plan} presents a graphical representation of the appointment intervals for a subset of clinics. In each subfigure, the horizontal axis represents the patient index, while the vertical axis represents the length of the appointment interval between two consecutive patients. 
It can be observed that the appointment intervals exhibit a ``Dome" pattern, in which the length of the appointment interval gradually increases at the beginning of an appointment session and gradually decreases towards its end.
This aligns with the conclusions on optimal appointment scheduling reported in the literature \cite{denton2003sequential, klassen1996scheduling, hassin2008scheduling}.
Shorter appointment intervals at the beginning of a session help minimize the doctor's idle time. As patients arrive, the intervals gradually increase to reduce their waiting time. Finally, towards the end of the session, the intervals shorten again to minimize the doctor's overtime. Therefore, this paper selects several ARs to explore how their semantics generate the ``Dome" pattern, as shown below. Their expression has been explicitly simplified for analysis. 

\begin{equation}
A_i = (i-CR)M+ (\min \{ 0.2CR \times i^2 , 0.8\} - 0.2) M
\end{equation}
\begin{equation}
A_i = (i-0.31003)M + 0.18 \min \{ PW(i-PW-1) / CV , 1\} iV
\end{equation}
\begin{equation}
A_i = iM+ 0.43479 \min \{ (0.43479i-2) \sqrt{i} , i \} V
\end{equation}

They are among the best AR for the ninth, tenth, and fourteenth clinics, respectively. They all share a common step size $M$. Notably, a negative offset in the step size can reduce the appointment intervals between patient 0 and patient 1. Furthermore, the correction terms of these rules also follow the same structure, incorporating a min function, exponents of $i$ greater than 1, and a subtraction of a constant. This results in their output being categorized into three types based on the patient index $i$. 
\begin{enumerate}
\item For small values of $i$, the term with exponents of $i$ greater than 1 is smaller than the constant offset in the subtraction, resulting in a negative value within the min function. This leads to shorter appointment intervals at the beginning of the appointment session. For example, under Rule (24), when $i < 4$, the condition $0.2CR \times i^2 <0.2$ holds (with $CR=0.1$). As a result, the correction term becomes negative, and the appointment intervals are shorter than the base step size. 
\item For medium values of $i$, the term with exponents of $i$ exceeds the constant offset in the subtraction, resulting in a positive value within the min function. Since these terms involve exponents of $i$ greater than 1, the appointment intervals increase as $i$ grows. For instance, under Rule (24), when $ 4 \leq i < 7$, we have $0.2CR \times i^2 > 0.2$ but less than $0.8$. Consequently, the appointment interval becomes longer than the base step size. 
\item For large values of $i$, the results of the min function are constant or depends on terms with exponents of $i$ equal to 1 or less. This causes the appointment intervals to decrease again and eventually stabilize toward the end of the session. In Rule (24), when $ i \geq 7$, the condition $0.2CR \times i^2 > 0.8$ holds, leading to appointment intervals becoming shorter again. 
\end{enumerate}
In contrast, traditional AR typically do not include terms with exponents of $i$ greater than 1. Instead, they employ a piecewise approach to create variations in appointment intervals. This structural difference in the evolved AR leads to  the formation of  a ``Dome" pattern and offers insight into  designing AR for more effective scheduling. 
 
\section{Conclusion}
\label{Conclusion}

The main goal of this paper is to develop a GP approach for the automatic evolution of appointment rules (AR) with both dimensional consistency and high performance. This goal has been successfully achieved by proposing a method called GPDR, which integrates a dimension repair procedure. This procedure optimizes the dimensional consistency of expression trees while minimizing structural changes, preserving their potentially effective pattern. 
Experimental results demonstrate that GPDR outperforms both traditional AR and the state-of-the-art DAGP methods across various clinic sizes and conditions, achieving lower objective values and no dimensional gaps. Further analysis reveals that GPDR is computationally efficient. Semantic analysis of the evolved AR provides insight into the design of AR. By combining the exponents of $i$ greater than 1 with min functions, evolved AR can generate a distinctive ``Dome" pattern, enabling near-optimal appointment scheduling.

Future work will consider other uncertainties in appointment scheduling, such as patient heterogeneity and unpunctuality. Furthermore, the problem-independent nature of GPDR and its dimension repair procedure suggests that  it can be readily adapted to address other challenges involving dimensional requirements. 

\bibliographystyle{IEEEtran}
\bibliography{main}

\end{document}